\documentclass[sigconf]{acmart}

\usepackage{booktabs} % For formal tables
\usepackage{hyperref}
\usepackage{url}
\usepackage{adjustbox}
\usepackage{makecell}

\setcopyright{rightsretained}
\acmDOI{10.475/123_4}

\acmISBN{123-4567-24-567/08/06}

\acmConference[KDD 2018]{ACM SIGKDD Conference}{Aug 2018}{London, UK}
\acmYear{2018}
\copyrightyear{2018}

\begin{document}
\title{Using Rule-Based Labels for Weak Supervised Learning}
\subtitle{A ChemNet for Transferable Chemical Property Prediction}

\author{Garrett B. Goh}
\affiliation{\institution{PNNL}}
\email{garrett.goh@pnnl.gov}

\author{Charles Siegel}
\affiliation{\institution{PNNL}}
\email{charles.siegel@pnnl.gov}

\author{Nathan Hodas}
\affiliation{\institution{PNNL}}
\email{nathan.hodas@pnnl.gov}

\author{Abhinav Vishnu}
\affiliation{\institution{PNNL}}
\email{abhinav.vishnu@pnnl.gov}

\renewcommand{\shortauthors}{Goh G.B. et al.}

\begin{abstract}
With access to large datasets, deep neural networks (DNN) have achieved human-level accuracy in image and speech recognition tasks. However, in chemistry, data is inherently small and fragmented. In this work, we develop an approach of using rule-based knowledge for training ChemNet, a transferable and generalizable deep neural network for chemical property prediction that learns in a weak-supervised manner from large unlabeled chemical databases. When coupled with transfer learning approaches to predict other smaller datasets for chemical properties that it was not originally trained on, we show that ChemNet's accuracy outperforms contemporary DNN models that were trained using conventional supervised learning. Furthermore, we demonstrate that the ChemNet pre-training approach is equally effective on both CNN (Chemception) and RNN (SMILES2vec) models, indicating that this approach is network architecture agnostic and is effective across multiple data modalities. Our results indicate a pre-trained ChemNet that incorporates chemistry domain knowledge, enables the development of generalizable neural networks for more accurate prediction of novel chemical properties.
\end{abstract}

\begin{CCSXML}
<ccs2012>
<concept>
<concept_id>10010147.10010257.10010258.10010262.10010277</concept_id>
<concept_desc>Computing methodologies~Transfer learning</concept_desc>
<concept_significance>500</concept_significance>
</concept>
<concept>
<concept_id>10010147.10010257.10010293.10010294</concept_id>
<concept_desc>Computing methodologies~Neural networks</concept_desc>
<concept_significance>500</concept_significance>
</concept>
<concept>
<concept_id>10010147.10010178.10010179</concept_id>
<concept_desc>Computing methodologies~Natural language processing</concept_desc>
<concept_significance>300</concept_significance>
</concept>
<concept>
<concept_id>10010147.10010178.10010224</concept_id>
<concept_desc>Computing methodologies~Computer vision</concept_desc>
<concept_significance>300</concept_significance>
</concept>
<concept>
<concept_id>10010405.10010432.10010436</concept_id>
<concept_desc>Applied computing~Chemistry</concept_desc>
<concept_significance>500</concept_significance>
</concept>
<concept>
<concept_id>10010405.10010444.10010087</concept_id>
<concept_desc>Applied computing~Computational biology</concept_desc>
<concept_significance>300</concept_significance>
</concept>
<concept>
<concept_id>10003752.10010070.10010071.10010289</concept_id>
<concept_desc>Theory of computation~Semi-supervised learning</concept_desc>
<concept_significance>300</concept_significance>
</concept>
</ccs2012>
\end{CCSXML}

\ccsdesc[500]{Computing methodologies~Transfer learning}
\ccsdesc[500]{Computing methodologies~Neural networks}
\ccsdesc[300]{Computing methodologies~Natural language processing}
\ccsdesc[300]{Computing methodologies~Computer vision}
\ccsdesc[500]{Applied computing~Chemistry}
\ccsdesc[300]{Applied computing~Computational biology}
\ccsdesc[300]{Theory of computation~Semi-supervised learning}

\keywords{Weak Supervised Learning, Transfer Learning, Computer Vision, Natural Language Processing, Cheminformatics, Bioinformatics}

\maketitle

\section{Introduction}
\label{sec:intro}

In the chemical sciences, designing chemicals with desired characteristics, such as a drug that interacts specifically with its intended target, or a material with specified physical performance ratings, is, despite decades of research, still largely driven by serendipity and chemical intuition. Over the decades, various machine learning (ML) algorithms have been developed to predict the activity or property of chemicals, using engineered features developed using domain knowledge. Recent work have also started using deep neural networks (DNN)~\cite{dahl2014, mayr2016, ramsundar2015, hughes2016}, that are on average, typically more accurate than traditional ML models~\cite{gawehn2016, goh2017r}.

\subsection{Limitations of Feature Engineering and Data Challenges}
Compared to modern deep learning research, the use of DNN models in chemistry relies heavily on engineered features. While such an approach is advantageous because it utilizes existing knowledge, using engineered features may limit the search space of potentially developable representations. This is exacerbated in situations in which engineered features are not appropriate or inadequate due to the lack of well-developed domain knowledge.

With the growth of chemical data~\cite{goh2017r}, it may be desirable to fully leverage representation learning, which will enable one to predict novel chemical properties for which little or no feature engineering research has been performed. In computer vision research, this is achieved by using raw data. For example, unaltered images are used as the input in various CNN models~\cite{szegedy2015,he2015}. In chemistry, DNN models that leverage representation learning from raw data are starting to emerge. For example, with minimal feature engineering, molecular graphs have been used to train DNN models~\cite{duvenaud2015, kearnes2016}. Other approaches use 2D or 3D images to train convolutional neural network (CNN) models~\cite{goh2017c1, goh2017c2, wallach2015}, or SMILES strings to train recurrent neural network (RNN) models~\cite{bjerrum2017, goh2017s}.

One factor that complicates representation learning is the limited amount of usable labeled data in chemistry, which is significantly smaller than that available in modern deep learning research. For example, having 100,000 labeled datapoints is considered a significant accomplishment in chemistry. In contrast, in computer vision research, datasets like ImageNet~\cite{russakovsky2015} that includes over a million images are typically the starting point. While sizable chemical databases like PubChem~\cite{kim2015} and ChEMBL~\cite{gaulton2011} do exist, their labels are skewed towards biomedical data, and such databases are only sparsely labelled, where each labeled data (i.e. measurements) are only available for a small subset (typically under 10\%) of the entire database. \textit{Therefore, the current state of labeled chemical data is small and fragmented, which reduces the effectiveness of representation learning when using conventional supervised training approaches.}

\subsection{Contributions}
Our work addresses the small and fragmented data landscape in chemistry. This is achieved by leveraging rule-based knowledge obtained from prior feature engineering research in chemistry~\cite{cherkasov2014} to perform weak supervised learning, and combining it with transfer learning methods used in modern deep learning research~\cite{oquab2014}. \textit{Specifically, we develop ChemNet, the first deep neural network that is pre-trained with chemistry-relevant representations, making it the analogous counterpart of a ResNet or GoogleNet for use in the chemical sciences.} Our contributions are as follows. 
\begin{itemize}

	\item We demonstrate that ChemNet learns chemistry-relevant internal representations, and when coupled with a transfer learning approach, can be used to predict novel chemical properties it was not originally trained on.
	\item We demonstrate the generalizability of ChemNet by predicting a broad range of chemical properties that are relevant to multiple chemical-affliated industries, including pharmaceuticals, biotechnology, materials and consumer goods.
	\item We demonstrate that the ChemNet pre-training approach is network architecture agnostic and effective across multiple data modalities, including both CNN and RNN models.
	\item We demonstrate that ChemNet-based models outperform otherwise identical models that are trained using conventional supervised learning, and that ChemNet also matches or exceeds the current state-of-the-art DNN models in the chemistry literature.
\end{itemize}

The organization for the rest of the paper is as follows. In section 2, we outline the motivations in developing a chemistry-relevant rule-based weak supervised learning approach, and the design principles behind ChemNet. In section 3, we examine the datasets, its broad applicability to chemical-affliated industries, as well as the training protocols used for pre-training ChemNet, and for evaluating its performance on unseen chemical tasks. Lastly, in section 4, we explore different ChemNet models, and the various factors that affect model accuracy and generalization. The best ChemNet model was then evaluated against other DNN models trained using conventional supervised learning approaches.

\subsection{Related Work}
Transfer learning is an established technique in deep learning research~\cite{oquab2014}. This approach first trains a neural network on a larger database, before fine-tuning it on a smaller dataset. For example, using ResNet that has been pre-trained on ImageNet to classify various common objects, may be used with transfer learning techniques to classify specific clothing type. In addition, as long as there is sufficient overlap in the "image space" on which the network was trained on, seemingly unrelated outcomes can be achieved. For example, a model pre-trained on ImageNet can also be fine-tuned to classify medical images~\cite{shie2015}. While medical applications are seemingly unrelated to conventional image recognition tasks, both sets of data are natural photographs (i.e. in the same "image space") and thus the lower-level basic representations can be utilized.

In the chemistry literature, because 2D molecular diagrams are substantially different from natural photographs, and chemistry-specific information is encoded into the image channels, existing pre-trained models in the computer vision literature on RGB images for example, would not be fully applicable. In addition, there are limited examples of using weak supervised learning to utilize large chemical databases in traing neural networks. \textit{Thus, the challenge of how to convert a sparsely labeled chemical database into a usable form that can be used in a transfer learning approach for existing chemistry DNN models is a non-trivial task. In our work, we will use molecular descriptors to generate consistent and inexpensive rule-based labels, combined with a weak supervised learning approach to train ChemNet to develop chemically-relevant internal representations, which is an approach that is conceptually unique relative to existing methods.}

\section{ChemNet Design}
\label{sec:design}

In this section, we provide a brief introduction to molecular descriptors, and its role as alternate labels for weak supervised learning of ChemNet. Then, we document the design principles behind ChemNet.

\begin{figure*}[!htbp]
\centering
\includegraphics[width=.8\paperwidth]{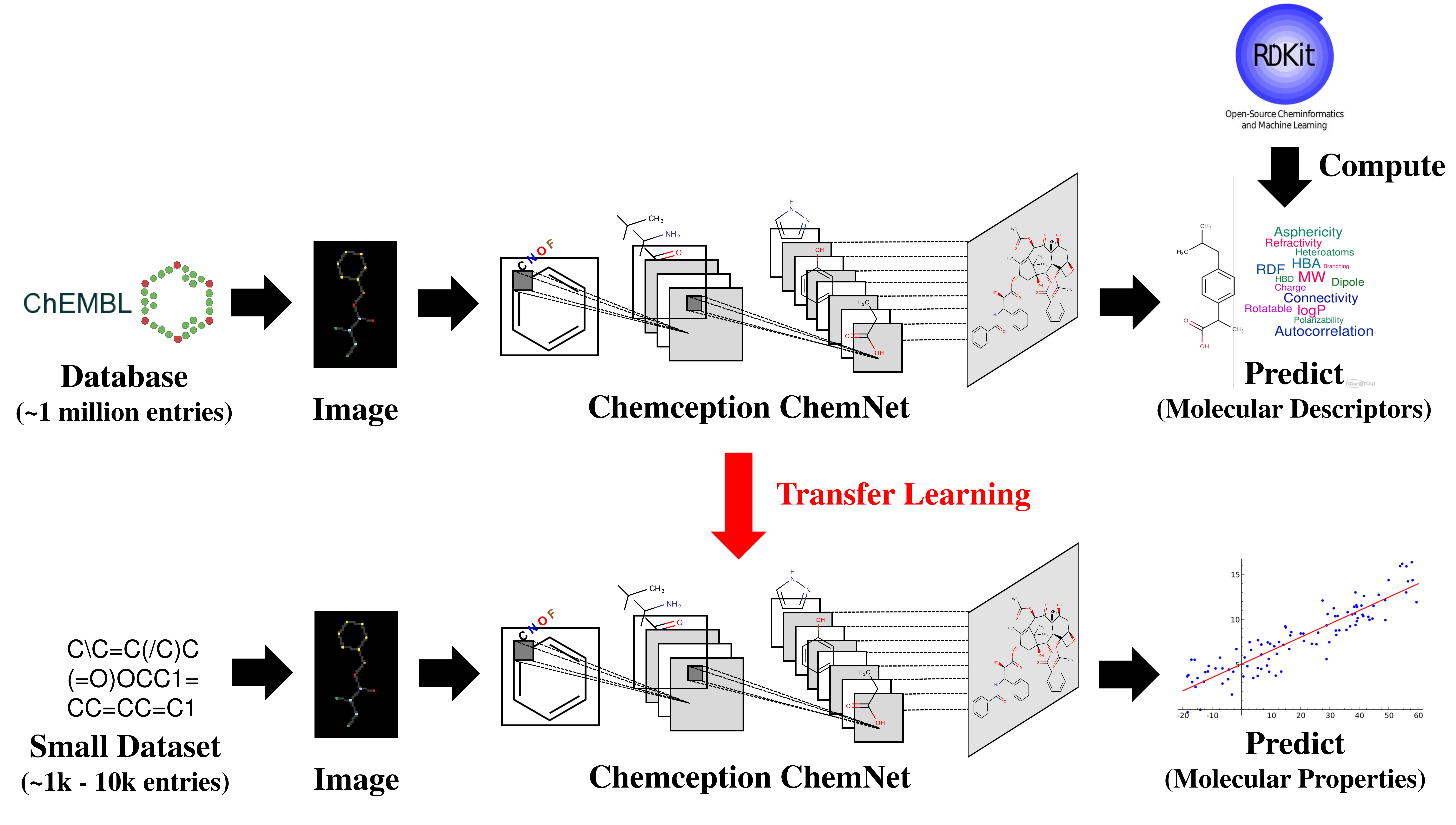}
\caption{\small Schematic illustration of ChemNet pre-training on the ChEMBL database using rule-based molecular descriptors, followed by fine-tuning on smaller labeled datasets on unseen chemical tasks.}
\label{fig:fig1}
\end{figure*}

\subsection{Alternate Labels for Weak Supervised Learning}
Much of the success in modern computer vision research comes from the availability of large labeled datasets like ImageNet~\cite{russakovsky2015}. However, in the chemical sciences, generating labels are both resource-intensive and time-intensive endeavors, and often the goal in chemistry research is to predict the chemicals with desired characteristics (i.e. the labels). Futhermore, because of the way deep neural networks are trained to recognize patterns, they cannot be easily programmed with specific rules for chemistry. Therefore, in order to develop a generalizable "chemistry-expert" DNN model that incorporates existing domain knowledge, we need to train a neural network in a manner such that it will learn basic and universal chemical representations. 

An analagous but hypothetical scenario that explains the data challenges in chemistry in a traditional computer vision research context would be as follows. In this fictitious example, the research objective is to classify object's names in ImageNet, but obtaining the names for a large dataset is unfeasible. However, concepts associated with images can be easily obtained. For example, in a classification of 3 images [cats, trees, cars], let us assume alternative labels that describe its texture [furry, rough, smooth] and whether it is man-made [natural, natural, artificial] can be easily computed. Therefore, when a CNN model is trained on these alternate labels in a weak supervised learning approach, representations that are developed to classify texture and whether an object is man-made, can be leveraged during a second transfer learning phase when the model is fine-tuned on smaller labeled dataset that has labels for the object's names. In this example, developing internal representations to identify a furry surface from images, may assist in identifying cats from trees and cars, as fur is a distinguishing characteristc that cats have. Similarly, developing other representations to identify artifical components such as wheels from the images, may assist in identifying cars. 

\subsection{Using Rule-Based Features to Teach Chemistry Representations}
Moving back to the chemical sciences, our approach is therefore to use molecular descriptors as alternate labels to perform weak supervised learning, after which transfer learning methods can be applied to fine-tune the pre-trained model directly on smaller labeled datasets of other unseen chemical properties of interests. Molecular descriptors are engineered features developed through historical research that stretches back to the late 1940s~\cite{platt1947}, and over 5000 molecular descriptors have been developed from rule-based chemistry knowledge~\cite{todeschini2008}. Molecular descriptors are typically computable properties or rule-based descriptions of a chemical's structure. Some molecular descriptors, such as hydrogen bond donor count, correspond to intuitive chemical knowledge that chemists use to conceptualize and understand more complex chemical phenomena. On other hand, there are descriptors such as the Balaban's J index~\cite{balaban1982} that may not be as intuitive, but nevertheless is an abstract topological description of a chemical that has been useful in various modeling studies.

In the absence of copious amount of data, the representation learning ability of deep neural networks may not learn optimal features. Our solution as illustrated in Figure~\ref{fig:fig1} uses molecular descriptors to generate consistent and inexpensive rule-based labels for large chemical databases that are typically sparsely and inconsistently labeled. Rule-based labels are then used to train ChemNet in a supervised manner, using a multi-task learning configuration, where the network attempts to predict all molecular descriptors simultaneously. In the process, the neural network develops "chemistry-relevant" representations that we hypothesize will serve as a better initialization of the network's weights when fine-tuning to other smaller and unrelated chemical tasks to be predicted.

\subsection{Hypothesis Behind ChemNet}
In this work, we first evaluated the ChemNet approach, on the Chemception CNN-based model~\cite{goh2017c1,goh2017c2}. As there is a strong relationship between a chemical's structure and its property, we hypothesize that the use of images of molecular drawings with a CNN-based model will therefore help facilitate the learning of chemistry-relevant structural representations. Then, we evaulated the ChemNet approach on SMILES2vec~\cite{goh2017s}, an RNN-based model that directly uses SMILES strings for predicting chemical properties. SMILES is a "chemical language" ~\cite{weininger1988} that encodes structural information into a compact text representation. As there is a one-to-one mapping between specific characters in the string to specific structural elements of the chemical, we hypothesize that ChemNet will also facilitate the learning of chemistry-relevant structural representations even when using a text representation of the chemical. Lastly, by using multi-task learning, we anticipate that the shared representation learned will be more generalizable, and it will be used as building blocks to develop more sophisticated and task-specific representations when fine-tuning on smaller datasets.

\section{Methods}
\label{sec:background}

In this section, we provide details on the datasets used, data splitting and data preparation steps. Then, we document the training and transfer learning protocols, as well as the evaluation metrics used in this work.

\subsection{Dataset for Pre-Training}
ChemNet was first trained on the ChEMBL~\cite{gaulton2011} database, which is a manually curated database of bioactive molecules with drug-like properties. In this work, after curation, approximately \textasciitilde1,700,000 compounds were used. In the initial pre-training stage, we compute molecular descriptors that serve as inexpensive and consistent (i.e. no missing) rule-based labels. Specifically, we used RDKit~\cite{landrum2016} to compute a list of \textasciitilde100 2D descriptors that includes basic computable properties (e.g MW, logP, etc.), connectivity, constitutional and topological descriptors.

\subsection{Dataset for Performance Evaluation}

Once ChemNet has been pre-trained, we fine-tune and evaluate ChemNet's performance on smaller datasets. To ensure that our results are comparable with contemporary DNN models reported in the literature~\cite{wu2017} and earlier work on Chemception CNN models~\cite{goh2017c1, goh2017c2} and SMILES2vec RNN models~\cite{goh2017s}, we used the Tox21, HIV, and FreeSolv dataset from the MoleculeNet benchmark~\cite{wu2017} for predicting toxicity, activity and solvation free energy respectively. The datasets used (Table~\ref{table:1}), comprises of a mix of large vs small datasets, physical vs non-physical properties and regression vs classification problems. None of the above-mentioned chemical tasks are related to the molecular descriptors used to train ChemNet, and thus also serve as a measure of ChemNet's ability to generalize to predict unseen chemical properties.

\subsection{Industrial Application}
Based on the datasets tested, we elaborate on applications to the chemical industry. First, toxicity prediction is of high relevance, most notably for chemicals that require FDA approval, which includes drugs and other therapeutics (pharmaceuticals) as well as cosmetics (consumer goods).~\cite{kruhlak2007} Activity prediction is proxy to how well-suited a chemical may be as a drug, and therefore is of relevance to both pharmaceuticals and biotechnology industries.~\cite{buchwald2002}  Solvation free energy values are computable by physics-based simulations, and such methods are currently being employed by pharmaceuticals, consumer goods and materials industries.~\cite{chodera2011} Therefore, using neural networks to predict such computable properties with similar accuracy will potentially lead to several orders of magnitude speed up compared to traditional computational chemistry simulations that typically take on the order of minutes to hours for each calculation.

\subsection{Data Preparation}
The preparation of chemical image data is identical to that reported by earlier work~\cite{goh2017c1}. Briefly, SMILES strings are converted to their respective 2D molecular structures using RDKit~\cite{landrum2016}. The coordinates of the molecule is then used to map it onto a discretized image of 80 x 80 pixels that corresponds to 0.5 A resolution per pixel. We initially used the greyscale "color-coding" scheme reported in the earlier Chemception paper~\cite{goh2017c1}. However, our subsequent experiments utilized the more sophisticated 4-channel "color-coding" scheme, where each atom and bond pixel is assigned a "color" based on its local (i.e. pixel-specific) atomic/bond properties, such as atomic number, partial charge, valence and hybridization. Specifically, we used the "EngD" augmented image representation "color-coding", and further details about this data preparation protocol can be obtained from published work~\cite{goh2017c2}.
 
The preparation of the chemical text data is identical to that reported by earlier work~\cite{goh2017s}. The SMILES string were first canonicalized using RDKit~\cite{landrum2016}, then unique characters in the string were then mapped to one-hot vectors. Zero padding was also applied to both the left and right of the string to construct uniform entries that were 250 characters long.

\begin{table}[!t] 
		\begin{center}
		\begin{tabular}{|c|c|c|c|}
				\hline
				Dataset & Property & Task & Size \\
				\hline\hline
				Tox21 & \makecell{Non-Physical \\(Toxicity)} & \makecell{Multi-task \\ classification} & 8014 \\
				\hline
				HIV & \makecell{Non-Physical \\(Activity)} & \makecell{Single-task \\ classification} & 41,193 \\
				\hline
				FreeSolv & \makecell{Physical \\(Solvation)} & \makecell{Single-task \\ regression} & 643\\
				\hline
		\end{tabular} 
		\end{center}
\caption{Characteristics of the 3 datasets used to evaluate the performance of ChemNet.}
\label{table:1}
\end{table}

\subsection{Data Splitting}
The dataset splitting steps are identical to that reported previously~\cite{goh2017c1}. We used a 5-fold cross validation protocol for training and evaluated the performance and early stopping criterion of the model using the validation set. We also included the performance on a separate test set as an indicator of generalizability. Specifically, for the ChEMBL database, Tox21 and HIV dataset, 1/6th of the dataset was separated out to form the test set, and for the Freesolv dataset, 1/10th of the dataset was used to form the test set. The remaining 5/6th or 9/10th of the dataset was then used in the random 5-fold cross validation approach for training ChemNet.

For classification tasks (Tox21, HIV), we also over-sampled the minority class to address the class imbalance observed in the dataset. This was achieved by computing the imbalance ratio and appending additional data from the minority class by that ratio. The oversampling step was performed after stratification, to ensure that the same molecule is not repeated across training/validation/test sets.

\subsection{Training the Neural Network}
ChemNet was trained using a Tensorflow backend~\cite{abadi2016} with GPU acceleration using NVIDIA CuDNN libraries\cite{chetlur2014}. The network was created and executed using the Keras 2.0 functional API interface~\cite{chollet2015}. We use the RMSprop algorithm~\cite{hinton2012} to train for 50 epochs for ChEMBL, or 500 epochs for Tox21, HIV, FreeSolv, using the standard settings recommended (learning rate = $10^{-3}$, $\rho = 0.9$, $\epsilon = 10^{-8}$). We used a batch size of 32, and also included an early stopping protocol to reduce overfitting. This was done by monitoring the loss of the validation set, and if there was no improvement in the validation loss after 10 (ChEMBL) or 50 (Tox21, HIV, FreeSolv) epochs, the last best model was saved as the final model. In addition, for images, during the training of the ChemNet, we performed additional real-time data augmentation to the image using the ImageDataGenerator function in the Keras API, where each image was randomly rotated between 0 to 180 degrees.

Unless specified otherwise, we used the weights of the final ChemNet model trained on molecular descriptors, as the initial weights to initialize subsequent individual models for predicting toxicity, activity and solvation energy. We also explored different fine-tuning protocols (see Experiments), where segments (i.e. a collection of convolutional layers) of ChemNet had its weights fixed.

\subsection{Loss Functions and Performance Metrics}
For classification tasks (Tox21, HIV) we used the binary crossentropy loss function, and for regression tasks (ChEMBL, FreeSolv) we used the mean-squared-error loss function. For the initial ChemNet pre-training on the ChEMBL database, we performed min-max normalization on the molecular descriptors, and these normalized labels were used for training the neural network. This ensures that each molecular descriptor is given equal emphasis during training.

For classification tasks (Tox21, HIV), the evaluation metric reported in our paper that determines model's performance is the area under the ROC-curve (AUC). For the FreeSolv dataset, the evaluation metric is RMSE. The reported results in the paper are the mean value of the evaluation metric, obtained from the 5 runs in the 5-fold cross validation.

\section{Experiments}
\label{sec:exp} 

In this section, we first conduct several experiments to determine the factors that may affect the performance and generalizability of ChemNet CNN models. Next, we demonstrated that the ChemNet approach on other data modalities by training a ChemNet RNN model. After establishing the best ChemNet model, we compare its performance against earlier Chemception/SMILES2vec models and other contemporary DNN models in the literature. 

\subsection{ChemNet Model Exploration}
In the absence of more data, network architecture has been a key driver in increasing model accuracy~\cite{szegedy2015,he2015}. Therefore, we first examine the network architecture and hyperparameters, followed by an evaluation of the image representation used. The full list of ChemNet CNN models explored is summarized in Table~\ref{table:2}.

We first evaluated the effect of Chemception architecture on the performance on the 3 chemical tasks: toxicity (Tox21), activity (HIV) and free energy of solvation (FreeSolv). From earlier work on optimizing the Chemception architecture, we evaluated both the baseline T1\_F32 and optimized T3\_F16 architectures~\cite{goh2017c1}. In the nomenclature used, T\textit{x} refers to the general depth of the network, and F\textit{x} refers to the number of filters in the convolutional layers. In addition, we also tested a wider and deeper Chemception T3\_F64 architecture.

\begin{table}[!t] 
		\begin{center}
		\begin{tabular}{|c|c|c|c|}
				\hline
				Model & Architecture & Image & Params \\
				\hline\hline
				T1\_F32\_std & T1\_F32 & Std & 276,603\\
				T3\_F16\_std & T3\_F16 & Std & 149,741\\
				T3\_F64\_std & T3\_F64 & Std & 2,369,681\\
				T1\_F32\_eng & T1\_F32 & Eng & 276,603\\
				T3\_F16\_eng & T3\_F16 & Eng & 149,741\\
				\hline
		\end{tabular} 
		\end{center}
\caption{Various pre-trained ChemNet models evaluated in this work investigated different network architectures and image representations.}
\label{table:2}
\end{table}

\begin{figure}[!htbp]
\centering
\includegraphics[scale=0.4]{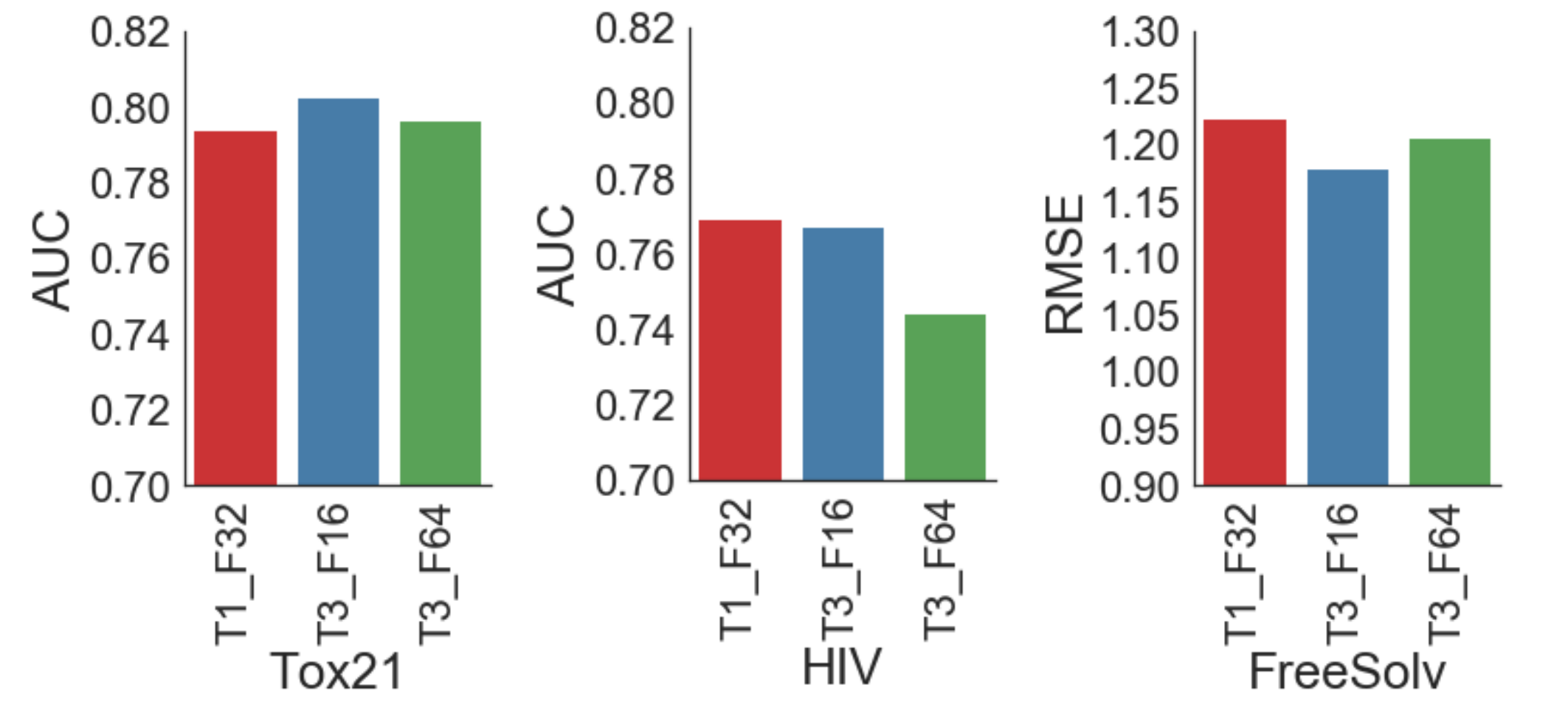}
\caption{\small The ChemNet T3\_F16 architecture generally had consistently better performance on the validation AUC/RMSE for toxicity, activity and solvation energy predictions. For Tox21 and HIV, higher AUC is better. For FreeSolv, lower RMSE is better.}
\label{fig:2}
\end{figure}
 
The T3\_F64 architecture with its wider layers that can accommodate more representations for the simultaneous prediction of \textasciitilde100 molecular descriptors attained the lowest normalized validation loss of $5.39 x 10^{-4}$ during ChemNet pre-training. This is slightly lower but still in the same order of magnitude as compared to that for T1\_F32 ($5.80 x 10^{-4}$) and T3\_F16 ($5.56 x 10^{-4}$) architectures. As illustrated in Figure 2, in terms of the validation metrics on the 3 unseen chemical tasks, the T3\_F16 architecture generally had consistently better performance than T1\_F32 even though it has approximately half the number of parameters. This implies that T3\_F16 is not suffering from underfitting. At the same time, the T3\_F16 architecture also has better performance than T3\_F64, which implies that adding more parameters while retaining similar network architecture of the same depth does not help to improve the generalizability of the model. \textit{Therefore, our findings indicate that amongst the network architectures tested in this work, a deep and narrow T3\_F16 architecture provides the best performance when generalizing to unseen chemical tasks.}

\subsection{Synergy of Image Representation and Transfer Learning}
\begin{figure}[!htbp]
\centering
\includegraphics[scale=0.33]{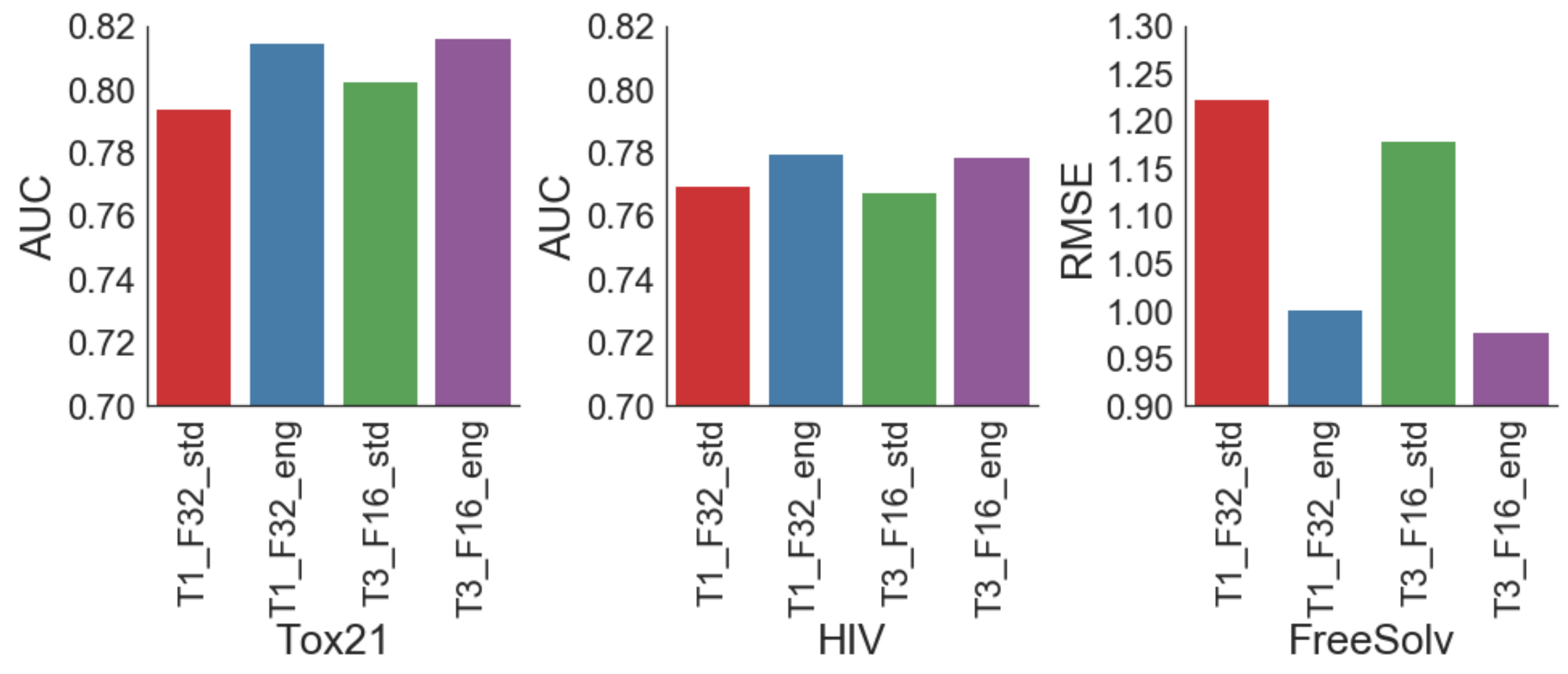}
\caption{\small Using augmented images results in consistently better performance on the validation AUC/RMSE for toxicity, activity and solvation energy predictions. For Tox21 and HIV, higher AUC is better. For FreeSolv, lower RMSE is better.}
\label{fig:3}
\end{figure}

Thus far, we have evaluated our results using the standard images reported in earlier work~\cite{goh2017c1}. However, subsequent improvements has shown that augmenting the image channels with basic atom and/or bond-specific chemical information improves overall performance~\cite{goh2017c2}. We note that the addition of localized (i.e. pixel-specific) chemical information to the image channels is complementary to ChemNet's transfer learning approach, as in this work, we are training the neural network to reproduce global chemical properties of the entire molecule (i.e. the entire image). As such, a combination of ChemNet with augmented images may lead to additional performance improvement. The results as summarized in Figure~\ref{fig:3} indicates that training ChemNet with augmented images consistently improved performance relative to standard images, and this is independent of the network architecture and the chemical task. \textit{Therefore, our findings indicate using augmented images of molecular drawings is a synergistic approach to ChemNet transfer learning methods.}

Having investigated various factors that may impact ChemNet performance, we come to the following conclusions: (i) using the Chemception T3\_F16 architecture provided the best consistent performance, and (ii) training with augmented images consistently improved performance. Therefore, for the remainder of this work, we will explore our results using the best model identified: T3\_F16\_eng.

\subsection{Transferability of Learned Representations}
The presented results in the preceding sections used the final weights of ChemNet as an initialization scheme for the individually trained networks for the smaller Tox21, HIV and FreeSolv datasets. We anticipate that due to the hierarchical nature of deep neural networks, it will learn hierarchical chemical representations, and more basic (lower-level) representations may not need to be re-trained. In order to determine which layers of ChemNet needs to be fine-tuned, we systematically explore the freezing of weights for various segments in the ChemNet model.

The T3\_F16 architecture is constructed from 12 segments, where each segment comprises of several convolutional layers that are grouped together based on similarities in their function. Specifically, ChemNet starts with a stem segment that has a single 4x4 convolutional layer, which is used to define the basic spatial region of the network. Following the stem segment is an alternating series of Inception-Resnet segments and Reduction segments. Each Inception-Resnet segment is a group of 4 convolutional layers that collectively perform inception-style operation with residual links, and the reduction segment is a group of 3 convolutional layers that downsamples the image. For further details of the network architecture, we refer our readers to earlier work~\cite{goh2017c1}.

Beginning with a ChemNet model that has all its weights frozen, we incrementally unfreeze (i.e. fine-tuned) the network starting with the top segment. We used this segment-based approach instead of a more conventional layer-based approach as the network architecture was designed with segments as the base unit in mind. The resulting model performance across all 3 chemical tasks were recorded as a function of number of segments fine-tuned. 

\begin{figure}[!htbp]
\centering
\includegraphics[scale=0.4]{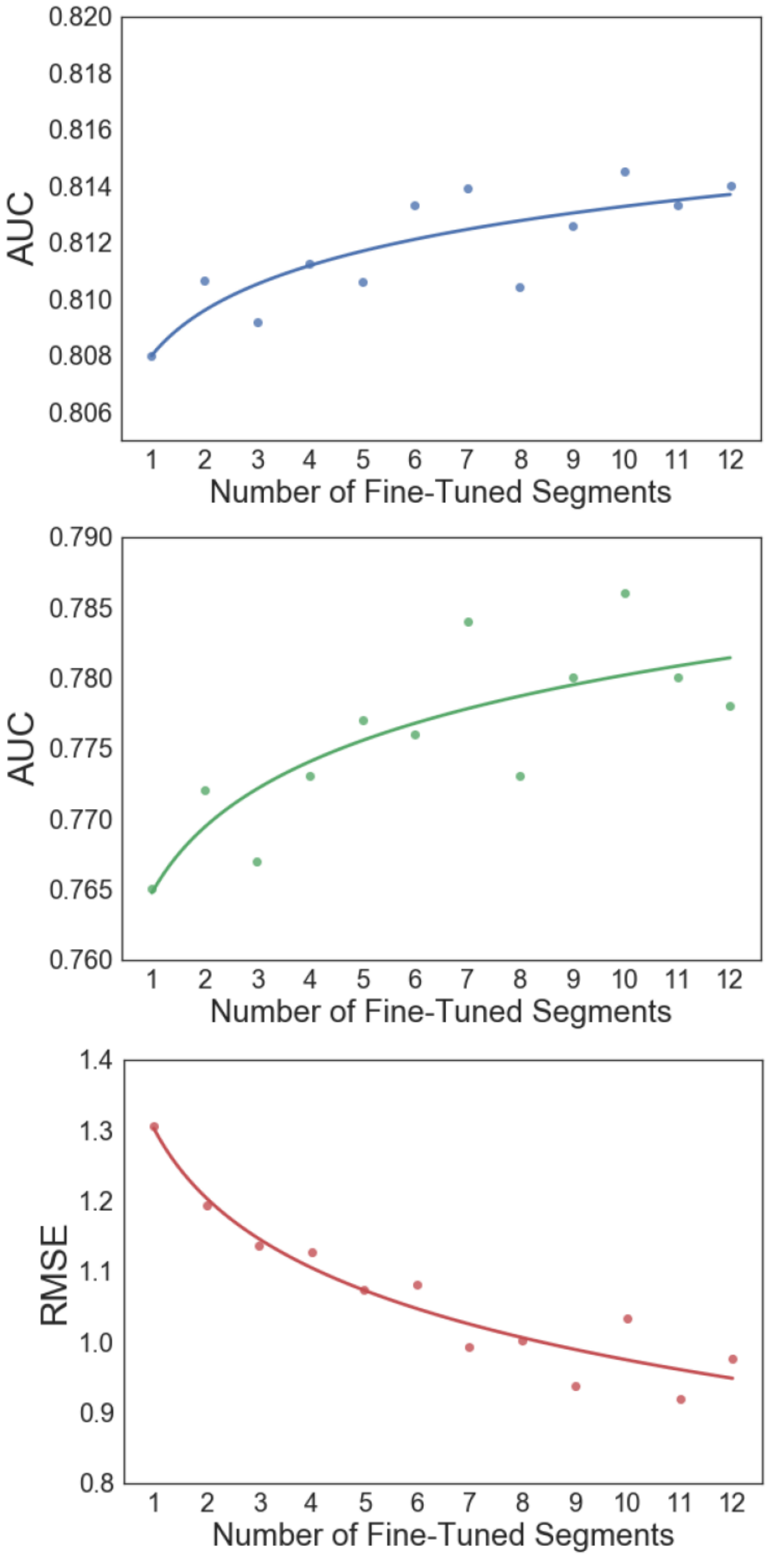}
\caption{\small Fine-tuning beyond the 7th segment of the ChemNet T3\_F16 architecture yield diminishing returns in performance improvement (validation AUC/RMSE) for toxicity, activity and solvation energy predictions.}
\label{fig:4}
\end{figure}

As illustrated in Figure~\ref{fig:4}, when less than 3 segments are fine-tuned, which is also when most of the network has its weights frozen, the model performance is relatively poor. This is an expected behavior as ChemNet was trained on a separate set of molecular descriptors that are unrelated to toxicity, activity and solvation energy predictions. As the number of segments that are fine-tuned increases, so does the model's performance. At the limit where all 12 segments are fine-tuned, we recover the results reported in the previous section.

We observed that there reaches a point of diminishing performance improvement at around the 7th segment, where additional fine-tuning of more segments do not consistently improve the results. This indicates that almost half of the network does not need to be re-trained, which suggest that the first half of ChemNet has developed basic chemical representations that are eminently transferable to other chemical tasks, and the second half of ChemNet develops more complex representations that needs to be fine-tuned for the specific property to be predicted. \textit{Therefore, our findings suggest that ChemNet, particularly for the lower layers have learned universal chemical representations that are generalizable to the prediction of other chemical properties.} 

\subsection{Performance Gain from ChemNet Transfer Learning}
\begin{figure}[!htbp]
\centering
\includegraphics[scale=0.32]{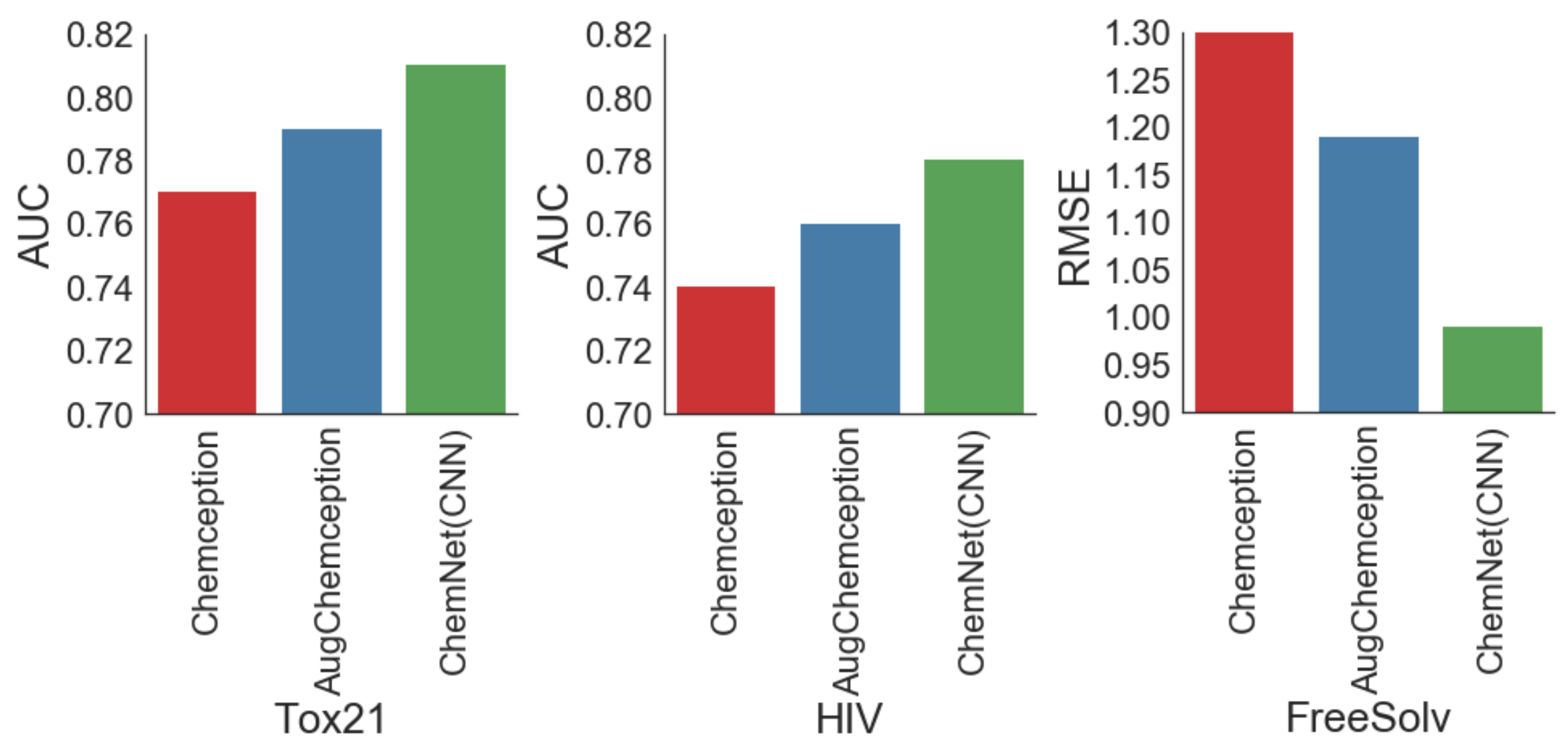}
\caption{\small ChemNet (based on the Chemception CNN model) provides consistently better performance on the validation AUC/RMSE for toxicity, activity and solvation energy predictions, as compared to earlier Chemception models that do not utilize transfer learning.}
\label{fig:5}
\end{figure}

Having identified the best ChemNet model and fine-tuning protocol, we now evaluate the performance of ChemNet against earlier Chemception models that do not utilize transfer learning. In Figure~\ref{fig:5}, we summarize the performance across various generations of Chemception-based models: Chemception refers to the original model that uses standard images~\cite{goh2017c1}, AugChemception refers to the modification of using augmented images~\cite{goh2017c2} and ChemNet refers to the results of this work.

Across all 3 chemical tasks, we observed that ChemNet achieves the best performance. Specifically, ChemNet achieves a validation AUC of 0.81 and 0.78 for toxicity and activity predictions respectively, and a validation RMSE of 0.99 kcal/mol for solvation free energy. Furthermore, we emphasize that there is no difference between AugChemception and ChemNet in terms of the network architecture and the images (data) used in the superivsed learning step, which means that the performance improvement is solely originating from the transfer learning techniques applied. \textit{Therefore, our findings indicate that the transfer learning techniques used in ChemNet provide a non-trivial improvement to model performance even when all other factors are held constant.}

\subsection{ChemNet on Other Data Modalities}

\begin{figure}[!htbp]
\centering
\includegraphics[scale=0.33]{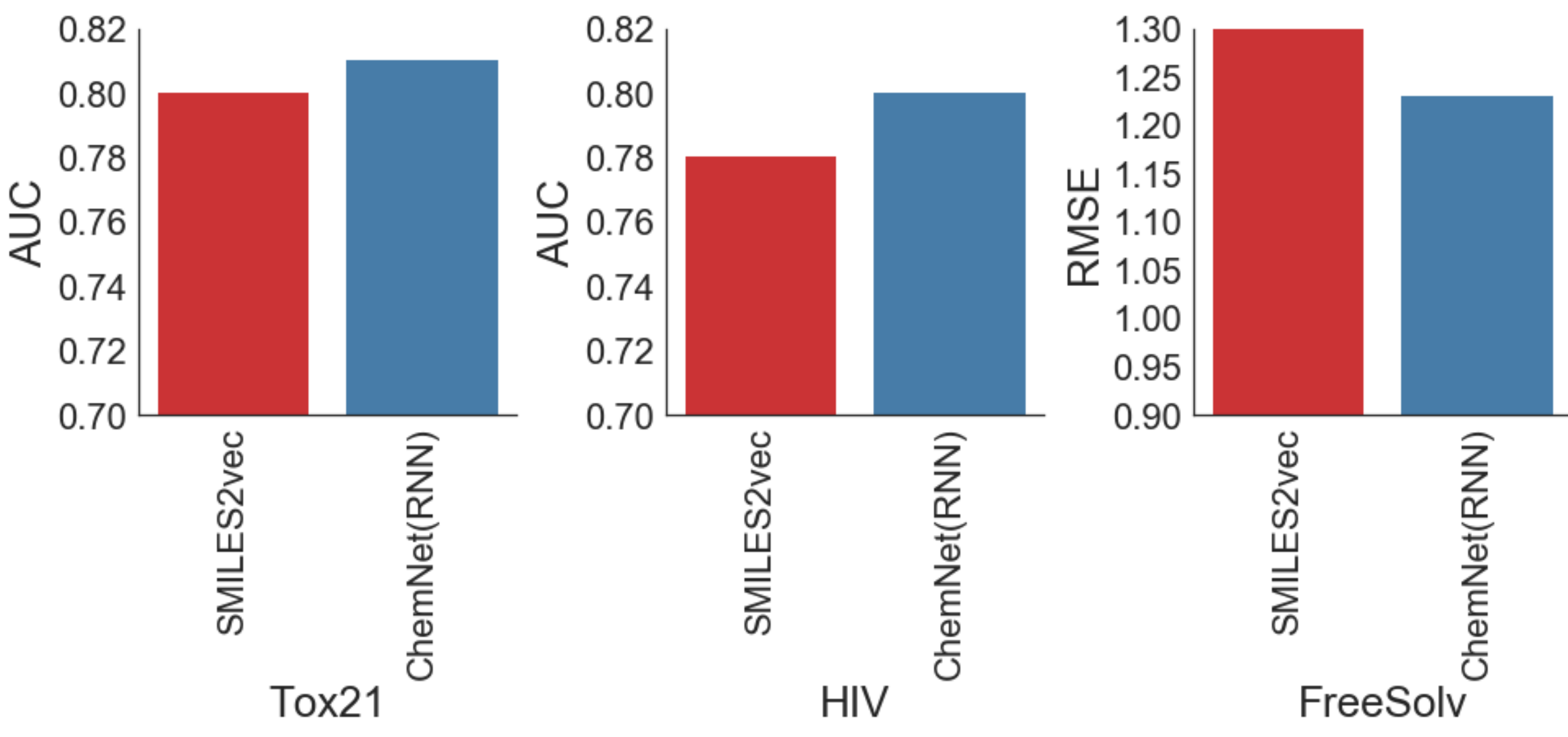}
\caption{\small ChemNet (based on the SMILES2vec RNN model) provides consistently better performance on the validation AUC/RMSE for toxicity, activity and solvation energy predictions, as compared to earlier SMILES2vec models that do not utilize transfer learning.}
\label{fig:6}
\end{figure}

Next, to demonstrate that the ChemNet approach is not unique to just CNN-based models and images, we pre-trained SMILES2vec, an RNN-based model that uses SMILES strings (chemical text) as input. The network architecture is based on the original SMILES2vec work~\cite{goh2017s}. The results as summarized in Figure~\ref{fig:6}, show that similar performance gain can be acheived with RNN models using a chemical text representation. Specifically ChemNet achieves a validation AUC of 0.81 and 0.80 for toxicity and activity predictions respectively, and a validation RMSE of 1.23 kcal/mol for solvation free energy. \textit{Therefore, our findings indicate that the ChemNet pre-training approach provides consistent performance improvement that is independent of the network's architecture and data modality.}

\subsection{ChemNet Against State-of-the-Art Models}
Having established that ChemNet provides consistently better performance than its counterpart Chemception and SMILES2vec models, we now perform benchmarks relative to other contemporary deep learning models in the literature. Specifically, we compare it to the MLP DNN model that was trained on molecular fingerprints~\cite{wu2017}. In addition, we also include the ConvGraph algorithm, which is a novel graph-based method for representing chemical data, and is the current state-of-the-art in many chemical tasks~\cite{wu2017}.

\begin{figure}[!htbp]
\centering
\includegraphics[scale=0.35]{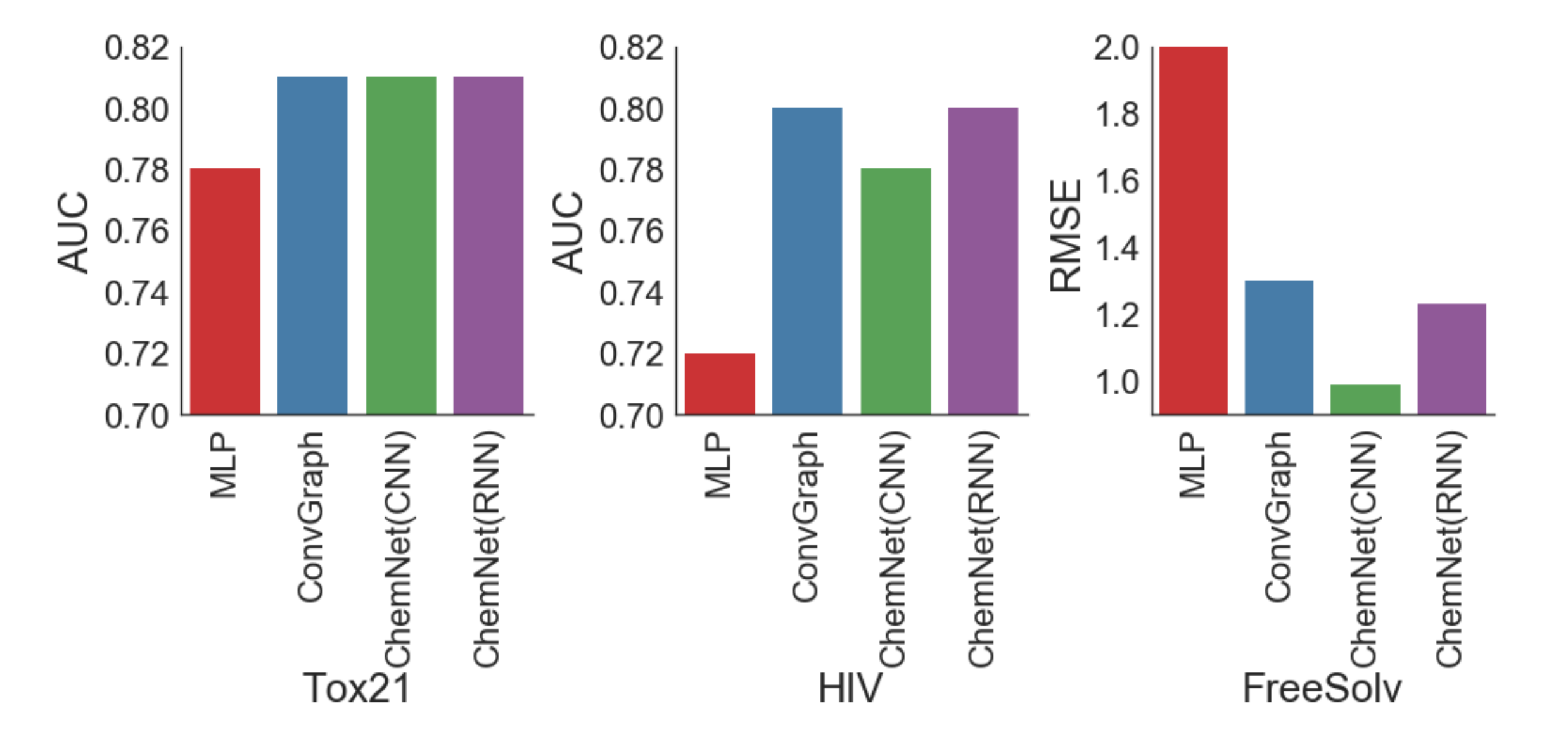}
\caption{\small ChemNet consistently outperforms MLP models trained on engineered features (molecular fingerprints), and matches the performance of ConvGraph on the validation AUC/RMSE for toxicity, activity and solvation energy predictions.}
\label{fig:7}
\end{figure}

As shown in Figure~\ref{fig:7}, compared to the MLP DNN model, which is the current state-of-the-art model that uses engineered features, we observe that both ChemNet(CNN) and ChemNet(RNN) consistently outperforms across all 3 chemical tasks. Relative to the ConvGraph algorithm, the best ChemNet model matches the performance for activity (val AUC 0.80 vs 0.80) and toxicity (val AUC 0.81 vs 0.81) predictions, and ChemNet significantly outperforms for solvation free energy (val RMSE 0.99 kcal/mol vs 1.30 kcal/mol) prediction.

\subsection{Rule-Based Weak Supervised Learning in Other Domains}
While the rule-based weak supervised learning approach that we have developed to train ChemNet is unique to chemistry, several design principles can be generalized to other domains. Specifically, the following factors were critical in enabling rule-based weak supervised learning in this work: (i) The availability of large datasets, but with the inability to generate ground-truth labels on a large-scale, and (ii) prior research in feature engineering and rule-based models which can be used to generate relatively inexpensive labels. Therefore, it is likely that other scientific, engineering and financial modeling applications, on which substantial research into rule-based models have been historically invested will benefit from this approach.

Furthermore, it should be emphasized that in this approach the process of pre-training the network is arguably more important than the accuracy of the initial ChemNet model in predicting the various rule-based labels (molecular descriptors). Technically, the rule-based labels on which ChemNet was pre-trained on are not related to the subsequent chemical properties that ChemNet was fine-tuned on. However, the hierarchical representations that deep neural networks form, are a good parallel to the hierarchical nature on which scientific concepts are built on top of one another. This suggests that the process of using rule-based transfer learning could potentially simulate the more conventional learning process of a domain expert, but without the need to explicitly introduce domain-specific rules. Thus, it is plausible that the lower layers of the network will learn representations that are analogous to simpler concepts in the technical domain, and such representations will give it the ability to adapt to different and/or unseen data. In the fine-tuning experiments in this work, which indicates that almost half of ChemNet has developed universal representations that can be re-used for predicting novel chemical properties, suggests that at least for the example of chemistry, such an approach is possible.

\section{Conclusions}
\label{sec:conclusions}

In conclusion, we have developed an approach for integrating rule-based knowledge with deep neural networks through weak supervised learning. Using the chemistry domain as an example, we demonstrate how rule-based knowledge (molecular descriptors), can be adapted with transfer learning techniques, to train CNN and RNN models on large unlabeled chemical databases of \textasciitilde1,700,000 chemicals. The resulting model, ChemNet, can be fine-tuned on much smaller datasets of \textasciitilde1000 to \textasciitilde10,000 chemicals to predict unrelated and novel chemical properties, that are of relevance to many chemistry-affliated industries. In addition, the ChemNet pre-training approach works effectively across network architectures and data modalities - where both CNN models (Chemception) using chemical images and RNN models (SMILES2vec) using chemical text had consistently better performance. For CNN models, we show that a combination of using augmented chemical images with the Chemception T3\_F16 architecture, and fine-tuning about half of the network provides the best and most generalizable performance, and RNN models also produce comparable results. ChemNet consistently outperforms all earlier versions of Chemception and SMILES2vec for toxicity, activity and solvation energy predictions, achieving a validation AUC of 0.81, 0.80 and validation RMSE of 0.99 kcal/mol respectively. In addition, ChemNet consistently outperforms contemporary deep learning models trained on engineered features like molecular fingerprints, and outperforms the current state-of-the-art ConvGraph algorithm for certain tasks. Furthermore, our fine-tuning experiments suggest that the lower layers of ChemNet have learned universal chemical representations inspired from rule-based knowledge, which improves its generalizability to the prediction of unseen chemical properties. Lastly, we anticipate the design principles behind our rule-based weak supervised learning approach will be adaptable to other scientific and engineering domains, where existing rule-based models can be used to generate data for training "domain-expert" neural networks in their specific field of application.

\begin{acks}
The authors would like to thank Dr. Nathan Baker for helpful discussions. This work is supported by the following PNNL LDRD programs: Pauling Postdoctoral Fellowship and Deep Learning for Scientific Discovery Agile Investment.
\end{acks}

\bibliographystyle{ACM-Reference-Format}
\bibliography{sample-bibliography}

\end{document}